\documentclass[10pt,twocolumn,letterpaper]{article}

\usepackage{wacv}
\usepackage{times}
\usepackage{epsfig}
\usepackage{graphicx}
\usepackage{amsmath}
\usepackage{amssymb}
\usepackage{float}

\newcommand{\norm}[1]{\left\lVert#1\right\rVert}

\wacvfinalcopy 

\ifwacvfinal
\def\assignedStartPage{1} 
\fi

\ifwacvfinal
\usepackage[breaklinks=true,bookmarks=false]{hyperref}
\else
\usepackage[pagebackref=true,breaklinks=true,colorlinks,bookmarks=false]{hyperref}
\fi

\ifwacvfinal
\else
\fi

\begin{document}

\title{Facial Keypoint Sequence Generation from Audio}

\author{Prateek Manocha, Prithwijit Guha\\
Department of Electronics and Electrical Engineering\\
Indian Institute of Technology Guwahati\\
{\tt\small prateek.manocha@iitg.ac.in, pguhau@iitg.ac.in}
}

\maketitle


\begin{abstract}
Whenever we speak, our voice is accompanied by facial movements and expressions. Several recent works have shown the synthesis of highly photo-realistic videos of talking faces, but they either require a source video to drive the target face or only generate videos with a fixed head pose. This lack of facial movement is because most of these works focus on the lip movement in sync with the audio while assuming the remaining facial keypoints’ fixed nature. To address this, a unique audio-keypoint dataset of over 150,000 videos at 224p and 25fps is introduced that relates the facial keypoint movement for the given audio. This dataset is then further used to train the model, Audio2Keypoint, a novel approach for synthesizing facial keypoint movement to go with the audio. Given a single image of the target person and an audio sequence (in any language), Audio2Keypoint generates a plausible keypoint movement sequence in sync with the input audio, conditioned on the input image to preserve the target person’s facial characteristics. To the best of our knowledge, this is the first work that proposes an audio-keypoint dataset and learns a model to output the plausible keypoint sequence to go with audio of any arbitrary length. Audio2Keypoint generalizes across unseen people with a different facial structure allowing us to generate the sequence with the voice from any source or even synthetic voices. Instead of learning a direct mapping from audio to video domain, this work aims to learn the audio-keypoint mapping that allows for in-plane and out-of-plane head rotations, while preserving the person’s identity using a Pose Invariant (PIV) Encoder.  
\end{abstract}

\begin{figure}[ht!]
\begin{center}
\centerline{\includegraphics[scale=1]{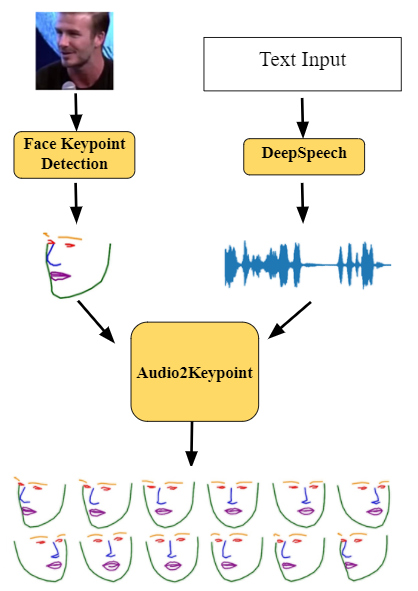}}
\end{center}
   \caption{{\bf Audio2Keypoint:} In this paper, a unique dataset, \emph{Vox-KP} has been introduced to learn the correlation between the audio and the movement of each of the facial keypoints. Further, Audio2Keypoint is proposed that generates plausible keypoint movement sequence (bottom) from a single target identity image (top left) and an audio or a text input (top right).}
\label{fig:first}
\end{figure}

\section{Introduction}
\label{sec:intro}

Recent years have witnessed an increasing attention in the area of deep generative models leading to interesting applications like deep-fakes \cite{chen2019hierarchical, vougioukas2018end, song2018talking, vougioukas2019realistic}, speech or music synthesis \cite{gansynth, arik2018neural, kumar2019melgan}, etc.
During conversations, human thoughts and expressions are conveyed through both voice and facial movements.

These movements add non-verbal information that helps the listener comprehend the speaker in a better way \cite{stacey2016contribution, cassell1999speech}, and also explains a person's preference for face-to-face conversations \cite{tarasuik2013seeing, o1993conversations}. 
This motivated the present work to develop a system for synthesis of facial keypoint sequence corresponding to an input audio segment and a reference face image. This keypoint sequence can be used further to generate photo-realistic videos of talking faces by deploying translation networks \cite{zakharov2019few, tang2019cycle}. Such a model can be deployed for a wide range of applications like video conferencing, video transmission in low bandwidth, virtual reality, chat bots, video dubbing, etc.

These applications require the generation of photo-realistic videos with lip movement in sync with the audio while preserving the person's identity. The synthesis of such photo-realistic talking faces is often challenged by the human observer's acuteness at identifying subtle abnormalities in the animated face image sequence (uncanny valley effect \cite{mori2012uncanny}) and its synchronization with the audio sequence. Another challenge is the incorporation of high geometric and photometric complexities of face models along with other attributes particular to an individual (hairstyle, clothing and accessories).

Facial reenactment is the process of re-animating a target video. Its success depends on the availability of source videos, which may not be always available. Existing works have proposed both video \cite{thies2016face2face, zollhofer2018state} and audio \cite{chung2017you, suwajanakorn2017synthesizing, vougioukas2018end} driven facial reenactment approaches. In contrast, methods proposed in \cite{chung2017you, vougioukas2018end} are helpful when there is a lack of driving videos. These models end up focusing mostly on preserving the facial characteristics and features and not much on the facial movement, which leads to a lack of video-realism. The focus of such methods on just the mouth region's movement results in an animated static image rather than a video. Suwajanakorn et al. \cite{suwajanakorn2017synthesizing} has shown photo-realistic video generation driven by audio for President Obama. However, such a method requires a person-specific dataset of roughly 17 hours, which limits its scalability and generalization to other people.

Several recent works such as \cite{thies2019neural, kim2019neural, song2020everybody} have shown the generation of highly realistic talking head videos, given an audio and a short video of the target identity. However, they primarily focus on realistic lip movement, neglecting facial expressions, and head motion, which often cause in-plane and out-of-plane head rotation. Also, these methods require short video of the target person, however, most of the time for greater generalization of the model, it is required to generate these videos from just a single image of the target identity, given speech as input. Although video-driven reenactment methods such as \cite{zakharov2019few, tang2019cycle} allow head motion, but these methods require a driving source video to transfer the facial keypoints on the target identity.

Most of these previous works treat the problems of facial motion and facial reconstruction together. However, to the best of our knowledge, for achieving this task, existing works have not explored the correlation between a person's speech (in any language) and facial movements (characterized by facial key points). Thus, to handle the above mentioned issues, this proposal explicitly focuses on the facial keypoint movement caused by speech production. These two modalities are strongly correlated \cite{yehia1998quantitative, nazzaro1970auditory}, but lie in different spaces. Thus, a dataset is required to model their correlation. This correlation allows us to consider the complete facial structure and preserve the same (for a target person) while generating the audio guided facial movement (keypoint sequence).
\newline

It is evident that most previous works have directly the mapping from audio to video domain. In contrast, this proposal learns the correlation between the audio sequence and the movement of each of the facial key points. Later, these can be used for the generation of photo-realistic videos. Focusing on each facial keypoint allows the generation of smooth and coherent facial keypoint movements, while taking into account both the facial structure and the head movement, even for images and audio outside the training dataset. However, the task of learning this mapping requires a large dataset of audio-keypoint videos and need to consider the different facial keypoint movements due to varying facial structures.
\newline

In this context, this work has contributed a new dataset \emph{Vox-KP}, that relates the audio with the facial keypoint movements for over $150,000$ videos recorded at $224p$ and $25fps$. The proposed model, \emph{Audio2Keypoint} generates facial keypoint sequence of a target person for his reference image input and an input speech segment. Specifically, given an audio segment and a reference image, the audio is encoded using an audio encoder. From the reference image, first the facial keypoints are extracted and then these are encode using a Pose Variant (PV) encoder and a Pose Invariant (PIV) encoder. These encoding are then passed to the decoder for the synthesis of the keypoint sequence movement. Adversarial learning \cite{mirza2014conditional} is used to prevent the convergence of output sequence to mean of the facial key points. This ensures a smooth output that is similar to real sequences.
\newline

The significant contributions of the present work are outlined as follows.

\begin{enumerate}
  \itemsep0em 
  \item {\bf Vox-KP}: A new dataset created using over 150,000 videos of over 6000 different identities from \cite {chung2018voxceleb2} to establish the relationship between the audio and the facial keypoint movement. 
  \item {\bf Audio2Keypoint}: Given an input image and an audio segment, Audio2Keypoint is a novel deep learning method to learn the correlation between the key point movement and the given audio segment, while taking into account the speakers facial structure and characteristics.
  \item {\bf PIV Encoder}: The PIV encoder is a novel model architecture trained along with the generator and discriminator that helps the generator to learn pose invariant information while acting as a discriminator at the same time to help the generator preserve the input person's identity and facial characteristics.
\end{enumerate}

The rest of the paper is organized in the following manner. Related work is described in Section~\ref{sec:relWork}. Dataset, \emph{Vox-KP} is described in section~\ref{sec:voxKP}. The proposed method is described in section~\ref{sec:propWork}. Finally, a quantitative and qualitative analysis is performed on various baseline models and the existing work in section~\ref{sec:experiment}.
\section{Related Work}
\label{sec:relWork}

This proposal (Audio2Keypoint) is related to facial reenactment approaches where an audio clip and a single image of the target identity are provided as input. These methods can be categorized into facial animation and facial reenactment. While facial animation focuses on the synthesis of expressions and lip movement applied to a predefined avatar, facial reenactment aims to generate photo-realistic videos of the target identity. This section briefly reviews the existing works on video-driven and audio-driven reenactment based approaches.

\subsection{Video Driven Reenactment}
\label{subsec:vidRenact}

These methods usually rely on reconstructing a target face using a source video comprising the intended speech content \cite{geng2018warp, wiles2018x2face, thies2016face2face, kim2018deep, nagano2018pagan}. Using either the facial landmarks \cite{geng2018warp, qian2019make} or expression parameters \cite{wiles2018x2face}, the target face is reenacted by using the parameters from the video of the source actor. A warp-guided generative model is used in \cite{geng2018warp} to transfer the landmarks from a source driving video to the target face. Expression parameters were used by estimating a driving vector comprising of desired facial expressions and head poses to drive the target face image \cite{wiles2018x2face}. While these methods produce frame-wise photo-realistic images, they suffer from unnatural movements due to insufficient temporal continuity. 

Model-based approaches \cite{thies2016face2face, kim2018deep, nagano2018pagan} leverage a parametric 3D head model. Using these parametric models, they disentangle the geometry, expression, and pose of the source model, which are copied to the target video. Face2Face \cite{thies2016face2face} transfers the source's expressions to the target face and uses static skin texture and a data-driven approach to synthesize the target mouth region.  The GANs are used in \cite{kim2018deep} to translate the target actor's synthetic rendering to a photo-realistic video frame for handling skin deformations. However, this approach is person-specific. Although, model-based techniques provide full control over target video, these methods suffer from audio-video misalignment \cite{thies2016face2face, kim2018deep, nagano2018pagan} leading to unrealistic results. Neural textures along with deferred neural renderer wre used to generate high-quality facial reenactments in \cite{thies2019deferred}. Neural style-preserving visual dubbing \cite{kim2019neural} preserves the source actor's style when transferring the facial expressions and can handle dynamic video backgrounds. Recently, \cite{zakharov2019few} have shown the generation of photo-realistic videos by transferring the source video's facial landmarks to the target face. However, it requires the source video to be from the target actor due to a lack of landmark adaption. Using landmarks from a video of different actors leads to a personality mismatch and unrealistic results.

\subsection{Audio Driven Reenactment}
\label{subsec:audRenact}

Audio-driven facial reenactment methods aim to generate photo-realistic videos that are synchronized with the input audio segment. This is particularly challenging because there exists no one-to-one mapping to associate phonemes with visemes, i.e., any given audio input can have different expressions associated with it. Several audio-driven facial reenactment techniques exist that try to animate the entire face such as \cite{chen2019hierarchical, chung2017you, song2018talking, vougioukas2018end, vougioukas2019realistic, zhou2019talking, zhu2018high}. While methods such as \cite{chen2019hierarchical, chung2017you, vougioukas2018end, vougioukas2019realistic, song2018talking, peng2017reconstruction} can work for any arbitrary target face, but only a few methods such as \cite{suwajanakorn2017synthesizing} generate photo-realistic full-frame images. Although earlier approaches like \cite{fan2015photo, suwajanakorn2017synthesizing} can produce photo-realistic results, they require a large amount of person-specific training data of the target subject and assumes source and target to have the same identity, which reduces their generalization to other target actors. For instance, Suwajanakornet et al. \cite{suwajanakorn2017synthesizing} require around 17 hours of President Obama’s speech as training data. In contrast, the present proposal only requires one single image of the target actor while considering the movement of all facial key points. Taylor et al. \cite{taylor2016audio} used a sliding-window neural network to map audio features to visual features that are encoded by an active appearance model (AAM) \cite{cootes1998active} parameters. 

Target independent subject methods such as \cite{chung2017you, vougioukas2018end, zhou2019talking} use an audio clip and a still face image as reference for video generation. However, since these methods aim to map the audio to the visual domain directly, the final result is an animation of the given image rather than a smooth, natural video. Chung et al. \cite{chung2017you} uses a joint embedding of the input face image and the audio in latent space and then uses an encoder-decoder CNN model to animate the mouth of the still image. Vougioukas et al. \cite{vougioukas2018end} used a similar method but with temporal GAN to produce the talking face with the audio clip and target actor image as input. The output is a more coherent animation, but it still lacked realism. Zhou et al. \cite{zhou2019talking} used both audio and video as input to learn joint audio-visual representation. 

Recently many GAN based methods \cite{chen2019hierarchical, vougioukas2018end, song2018talking, vougioukas2019realistic, zhou2019talking} have used a single image of the target identity and an audio clip for the generation of facial animation. Chen et al. \cite{chen2019hierarchical} first transfers the audio to facial landmarks and then generates video frames conditioned on these landmarks. Song et al. \cite{song2018talking} use a conditional recurrent adversarial network that uses both audio and image features in recurrent units. However, in talking face videos generated by these 2D methods \cite{chen2019hierarchical, song2018talking}, the head pose remains almost fixed during talking. Most of these methods focus on addressing the synchronization of the lip movement with the audio instead of the overall face’s movement to improve realism. In contrast, Audio2Keypoint focuses on the movement of each facial landmarks by learning the correlation between the landmarks and the corresponding audio segments.

\section{Dataset: Vox-KP}
\label{sec:voxKP}

An audio-keypoint dataset \emph{Vox-KP} is introduced in this work for learning the correlation between the movement of facial keypoints and audio segments. For the creation of this Vox-KP, around $150,000$ videos of $6,112$ celebrities (recorded at a frame resolution of $224p$ and frame rate of $25fps$) are used from the VoxCeleb2 \cite{chung2018voxceleb2} dataset. VoxCeleb2 provides us with an audio-keypoint dataset of over $6000$ celebrities with wide variety of facial geometry and languages. \\

\noindent
\textbf{Annotations:} For a given a video, the speaker's facial movement over time is represented as an array of 2D facial keypoints) using an off-the-shelf face alignment model \cite{bulat2017far}. For each video frame, all the $68$ keypoints corresponding to chin, nose, eyebrows, eyes, outer and inner lips are saved. For visualization purposes, the obtained keypoints are rasterized into an RGB image using a predefined set of colors to connect the keypoints belonging to the same parts of the face. \\

\noindent
\textbf{Pseudo Ground Truth:} State of art face alignment model \cite{bulat2017far} is used to obtain facial keypoints on each frame. This provides us with the pseudo ground truth annotations. 

This strategy is adopted from the work presented in \cite{ginosar2019learning}. 
The use of these pseudo-annotations allows the training on a much larger dataset and eliminates the need of manual annotations. 

Results for the model trained using these (obtained) annotations are shown in Figure~\ref{fig:compare-intra} and \ref{fig:compare}.

\begin{figure*}[ht]
\begin{center}
\centerline{\includegraphics[scale=0.41]{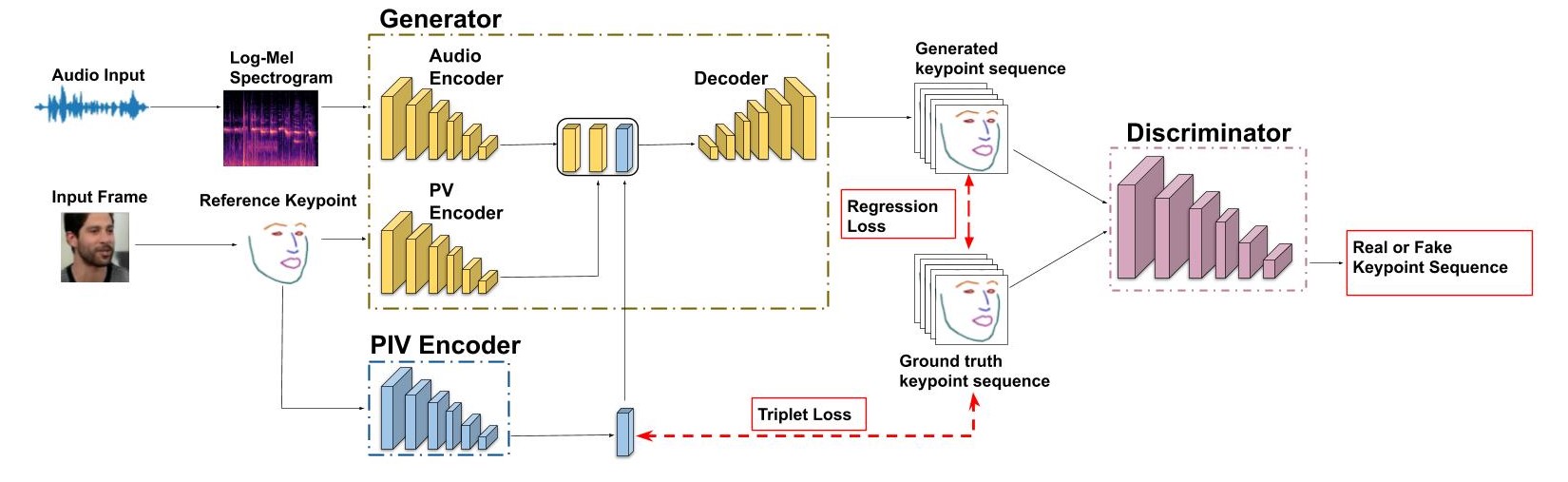}}
\end{center}
   \caption{Audio2Keypoint consists of three components. The input target frame and the audio segment first pass through the {\bf Generator}. The audio is first converted to spectogram and then encoded, while the reference keypoints are extracted from the input image and then passed to PV Encoder and {\bf PIV Encoder} separately. These encoding are then concatenated and passed to a decoder that generates the keypoint movement sequence, condition on the input image and the audio segment. The {\bf Discriminator} ensures that the generated keypoint sequences are real while being temporally smooth and coherent. Adversarial loss, along with regression and encoding loss, is used for optimization.}
\label{fig:model}
\end{figure*}

\section{Proposed Work}
\label{sec:propWork}

The proposed \emph{Audio2Keypoint} model aims to generate temporally smooth facial keypoint movement sequences which can be further used to synthesize photo-realistic videos of talking face of a target person. This involves the following inputs.

\begin{itemize}
  \item Audio segment (need not be of the same person)
  \item A reference image of the target person, which is also used as the starting frame of the generated video.
\end{itemize}
\noindent

The Audio2Keypoint model consists of the following three components. These are (a) Generator, (b) keypoint pose invariant (PIV) encoder, and (c) Discriminator. These are shown in Figure~\ref{fig:model} and are described in the following sub-sections.

\subsection{Generator}
\label{subsec:gen}

The 2D Log-Mel spectrogram, $s$ is first computed from the input audio segment $a$. The former is then fed to the generator's audio encoder. 

The 2D facial reference keypoints (more detail in Sub-section~\ref{subsec:propWorkImp}) are further fed to both Pose Variant (PV) and Pose invariant (PIV) encoders.\\

\noindent
The generator $G(s, k)$ is a fully convolutional network consisting of an audio encoder, pose variant encoder, and a decoder similar to the UNet \cite {isola2017image, ronneberger2015u} translation architecture. The generator's down-sampling path consists of audio and pose variant encoding. Using an architecture similar to UNet provides the up-sampling path/decoder with past and future temporal context, while the skip connections between the audio encoder and the decoder allow for high-frequency temporal information to flow through.\\

\noindent
{\bf PV Encoder} -- The information encoded by facial landmarks can be categorized into Pose Variant (PV) information, which depends on intrinsic Euler angles for the pose (yaw, pitch, roll) of a human face and pose invariant information. PV Encoder takes the landmark $k$ associated with the reference frame $r$ and gives its corresponding latent encoding. Since, $r$ is the first frame of the ground truth video corresponding to audio input $a$, the generator should also learn to begin the output keypoint sequence starting from $k$. 
\newline \newline
\noindent

{\bf Audio Encoder} -- Any plausible keypoint sequence should be temporally coherent and smooth, i.e., the displacement of each keypoint must be rational. This desired smoothness is attained by learning an audio encoding that represents the complete audio signal by taking into account the full temporal extent of the input $a$ rather than treating the audio signal recurrently. This is achieved by taking a 2D Log-Mel spectrogram of the input audio signal which is then down-sampled into a latent audio encoding using convolutional networks.
\newline \newline

\noindent {\bf Decoder} -- The decoder forms the up-sampling path of the generator with skip connections with the audio encoder. It concatenates audio encoder, PV encoder, and the PIV encoder and takes this concatenated vector as an input. The input is then up-sampled layer-by-layer using transposed convolution, which is then followed by convolution blocks to achieve the desired output keypoint sequence.

\subsection{Keypoint PIV Encoder}
\label{subsec:pivEnc}

A single encoder followed by a separate network for obtaining the PV and PIV information can be deployed following the proposals in \cite{peng2017reconstruction, yin2017multi}. However, such a method restricts the model's learning capacity by making it focus on a single encoding which may be sparse and not informative enough of both PV and PIV information for the input of 136-dimensional array $k$. Using separate encoders make the model focus on generation of compact PV and PIV encoding. \\
\newline
PIV encoder $E(k)$, takes $k$ as an input and gives the corresponding latent encoding. To learn pose invariant encoding, both the reference keypoint $k$ and all the facial key points present in the ground truth keypoint sequence $y$ must have similar latent encoding vector. To further boost the generator's training, we use triplet loss function, that uses the output keypoint sequence, ${\bf \hat y}$, as negative samples (refer Sub-section~\ref{subsec:lossFunc} for more details).

\subsection{Discriminator}
\label{subsec:disc}

Although the generator learns in a supervised manner using \emph{$L_1$} regression loss (Sub-section~\ref{subsec:lossFunc}), the generator may learn the mean of all the keypoint positions, thereby producing overly smooth keypoint movements with minimal lip movement. A discriminator \cite{chan2019everybody} is used to handle this issue and to distinguish between plausible facial keypoint sequence and fake sequences \cite{zakharov2019few,ginosar2019learning}.\\ 

The discriminator $D(y)$ consists of a convolution network that takes a plausible keypoint sequence (in the form of the temporal stack) as an input, calculates the displacement of each of the facial keypoint in the input sequence, and then predicts a single scalar to score the realism and the consistent nature of the input keypoint sequence. The discriminator takes the whole temporal stack of keypoint sequence as an input to ensure temporal coherence which in turn helps the generator learn to produce smooth and coherent facial keypoint sequences. The PatchGAN \cite{isola2017image} architecture is adapted for the discriminator network.

\subsection{Loss Functions}
\label{subsec:lossFunc}

During the training stage, the parameters of all three networks are trained simultaneously in an adversarial manner.  \\ 

\textbf{Generator} -- The parameters of the generator are optimized to minimize the overall loss function containing adversarial loss, regression loss and PIV encoding loss: \[L_{Gen} = L_{Adv} + L_{Reg} + L_{PIV-Gen}\]
The adversarial loss ($L_{Adv}$) corresponds to the realism score computed by the discriminator, which the generator ties to maximize to encourage the change in facial keypoints to be smooth and coherent. 
\[L_{Adv} = \norm{1 - D(\hat y)}_{L_{2}}\]
Regression loss ($L_{Reg}$), represents the $L1$ regression that allows the generator to learn to map the audio signal to the temporal stack of pose vectors $y$. 
\[L_{Reg} = \norm{y - \hat y}_{L_{1}}\]
The PIV encoding loss ($L_{PIV-Gen}$) forces the generator to produce output such that each facial keypoint frame in $\hat y$ encodes similar PIV information as that encoded by reference frame $k$.
\[L_{PIV-Gen} = \norm{e(k)- avg(e(\hat y))}_{L_{2}}\]

\textbf{PIV Encoder} -- The parameters of the PIV encoder are optimized to encode PIV information that eventually helps the generator preserve the characteristics of reference keypoints. The parameters are therefore optimized to minimize the following objective.
\[L_{PIV} = L_{Reg} + L_{Adv} + L_{PIV-PIV}\]
PIV Encoding loss of PIV Encoder ($L_{PIV-PIV}$) tries to learn PIV information from the input reference $k$ and ground truth frames $y$. A triplet loss function is employed for this. Specifically, we make the encoding vectors of the reference $k$ and the encoding of each frame in $y$ as close as possible while making the encoding of $k$ and $\bf \hat y$ as far apart in terms of root mean squared (RMS) distance. 

\[L_{PIV-PIV} = \norm{k-y} + \norm{k - \hat y} + \epsilon\]
 
 $L_{Reg}$ and $L_{Adv}$ makes PIV encoder act as a helper to the generator, encouraging it to match the ground truth performance, while $L_{PIV-PIV}$ helps it encode the PIV information with $\hat y$ acting as negative examples that makes the training faster and more stable by providing an additional constraint on the generator. 
 
\textbf{Discriminator} -- The discriminator tries to capture the spatial and temporal coherence. It tries to learn to classify whether a given sequence is real or not, thereby forcing the generator to learn to produce real-seeming facial keypoint sequences. The parameters of the discriminator are driven by the minimization of
\[L_{D} = \norm{1 - D(y)} + \norm{D(\hat y)}\]

The objective is to update the discriminator parameters to increase the realism score on ground truth sequence $y$ to $+1$ and decrease the score of synthesized keypoint sequence $\hat y$ to $0$.

\subsection{Implementation}
\label{subsec:propWorkImp}

From each keypoint sequence, we treat the first nose keypoint to be the base point, which is then subtracted from all the keypoints in the pseudo ground truth facial landmarks. This ensures that all training keypoint sequences have the same base point$(0,0)$. This ensures translation invariance. All keypoint sequences are then normalized using mean and standard deviation that are computed using first $10k$ sequences of the dataset. 

During training, audio of roughly $2.5$ seconds (corresponding to $64$ frames at $25fps$) is taken as input and is used to predict corresponding $68$ facial keypoint positions for each of the $64$ frames. However, no such restrictions are imposed during the testing and the network can operate for arbitrary audio segment duration.

\section{Experiments}
\label{sec:experiment}

Audio2Keypoint has several use cases such as audio-driven video synthesis for video conferencing and video dubbing. Given the target identity image and audio input, the generated keypoint sequence can be combined with facial landmark driven video synthesis methods \cite{zakharov2019few}. We first demonstrate the experimental setup and then perform a quantitative and qualitative evaluation.

\subsection{Experimental Setup}
\label{subsec:setup}

Audio2Keypoint has been trained using Tensorflow and takes approximately two days to train on 100,000 videos using an Nvidia V100 GPU. We optimize using Adam optimizer \cite{kingma2014adam} with a batch size of 32, along with a learning rate of $10^{-4}$ for all the three components. The model is trained for 300 epochs, and we get the lowest validation error at 260 epoch.

\begin{figure*}[ht]
\begin{center}
\centerline{\includegraphics[scale=0.65]{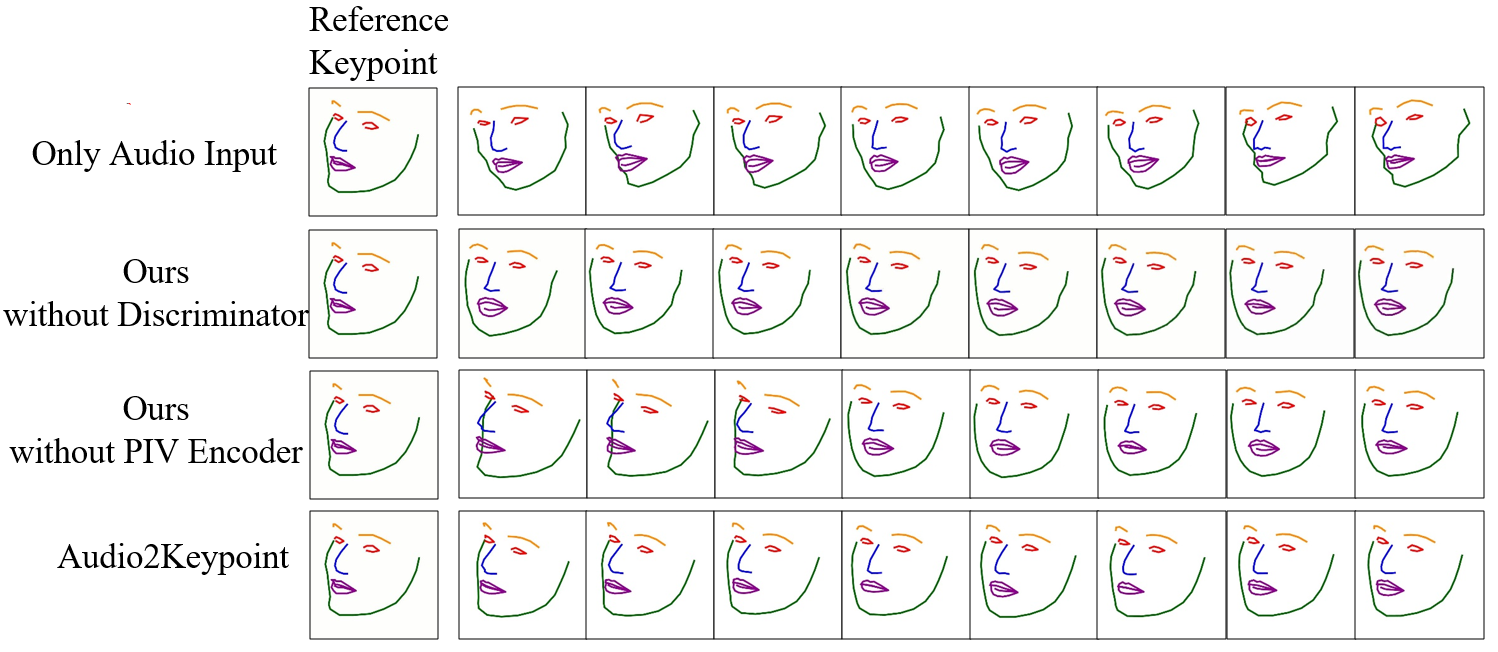}}
\end{center}
\caption{Comparison of Audio2Keypoint with various baseline models. In the absence of image input, the model finds it hard to learn the facial geometry and ends up learning a generic face irrespective of the target identity. Without the discriminator, the model learns to synthesize the reference keypoints, but all the facial keypoints remain static for most of the time. Audio2Keypoint, without the use of PIV Encoder performs equally good as Audio2Keypoint. However, it fails to learn the correct starting frame and deviates slightly from the reference facial structure. Finally, Audio2Keypoint overcomes all these shortcomings and generates a smooth and naturally looking keypoint movement sequence, while preserving the target person's facial structure.}
\label{fig:compare-intra}
\end{figure*}

\subsection{Ablation Study}
\label{subsec:ablation}

As shown in Figure~\ref{fig:model}, Audio2Keypoint consists of three components -- Generator, PIV Encoder, and Discriminator. We have performed an ablation study to evaluate the importance of each of these components. We use the following baselines for comparison.
\newline

\begin{figure*}[ht]
\begin{center}
\centerline{\includegraphics[scale=0.85]{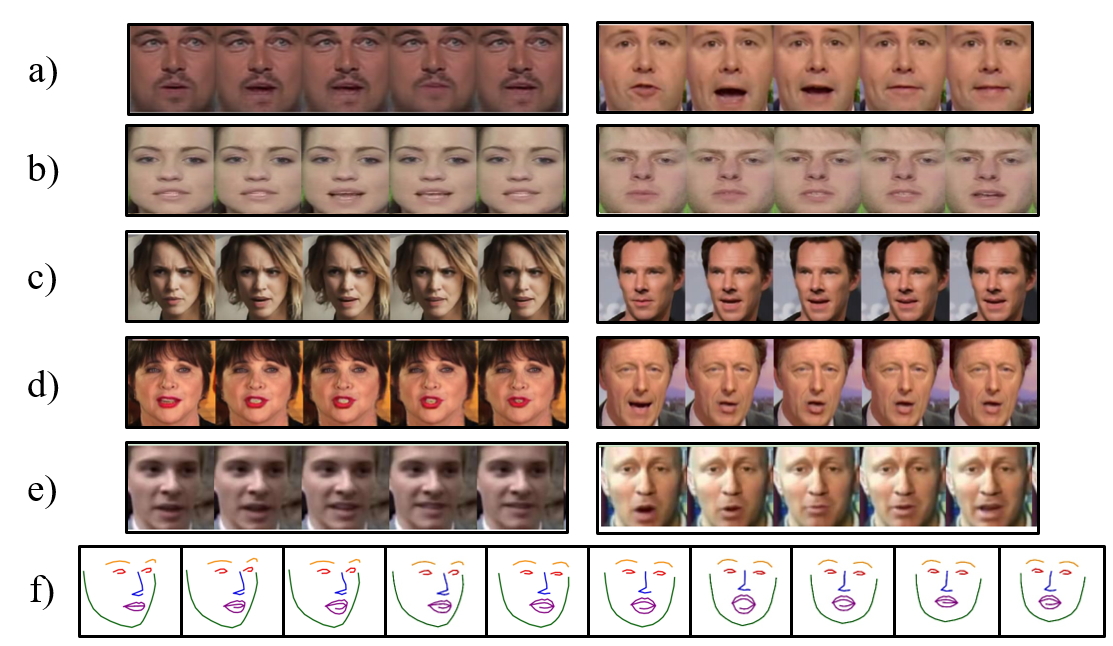}}
\end{center}
\caption{Comparison of Audio2Keypoint with various target independent audio-driven reenactment methods.   a) Chung et al. \cite{chung2017you} b) Vougioukas et al. \cite{vougioukas2018end} c) Zhou et al. \cite{zhou2019talking} d) Song et al. \cite{song2018talking} e) Chen at al. \cite{chen2019hierarchical} f) Audio2Keypoint. As compared to these methods, Audio2Keypoint focus on movement of each of the keypoints and thus leads to more smooth and natural movement of the landmarks. Rather than focusing on complete image generation, Audio2Keypoint focus explicitly on the facial keypoints movements which can then be used for generation of photo-realistic images guided by these keypoints.}
\label{fig:compare}
\end{figure*}

\noindent
{\bf Only Audio Input} -- This baseline model predicts the facial keypoint sequence given just the audio input. This model is similar to \emph{speech2gesture} \cite{ginosar2019learning} but is not person-specific. Absence of a target identity facial keypoint makes it hard for the model to learn the relative movement of each keypoint as there is no starting reference available. Hence, it learns a fixed starting frame as shown in Figure~\ref{fig:compare}.

\noindent
{\bf Audio2Keypoint, without PIV encoder} -- Given an audio and target identity image input, we use only the Generator and the Discriminator to synthesize the keypoint sequences. However, the model gives a smooth and natural-looking movement but fails to preserve the target identity's facial structure.

\noindent
{\bf Audio2Keypoint, without discriminator}: We compare Audio2Keypoint to predict the facial keypoint movement using just the translation architecture alone, without the adversarial discriminator. Due to the absence of the discriminator, the model fails to learn temporally smooth keypoints movements, leading to static facial keypoints, as shown in Figure~\ref{fig:compare}.

\noindent
{\bf Audio2Keypoint, the final model}: Finally, we compare all of these models with Audio2Keypoint, which uses all the three components to synthesize the facial keypoint movement sequence.

\subsection{Evaluation Metrics}
\label{subsec:metrics}

To evaluate the performance of Audio2Keypoint, we mainly use L1 regression loss on the validation dataset for each of the trained models. Apart from L1 regression loss, we also use percent of correct keypoints predicted (PCK) \cite{yang2012articulated}, a metric used mainly for pose detection problems. In PCK, a predicted keypoint is considered correct if the distance between the predicted and the ground truth lies within a certain threshold. We use $\alpha max(h, w)$ as the fixed threshold for all the key points, where $h$ and $w$ are the bounding box's height and width for that particular sequence. Unlike L1 loss, which penalizes a keypoint based on its distance from the ground truth, PCK has a more hard threshold as a keypoint's contribution falls to zero outside this threshold. Thus, L1 serves as a more suitable measure for our use case as we do not intend to predict the ground truth using Audio2Keypoint, but rather we are using it as a training signal to learn the correlation as there can be several plausible keypoint movements for the same speech utterance. We use $\alpha = 0.02$.

\subsection{Quantitative Analysis}
\label{subsec:quantitative}

We measure Audio2Keypoint performance using the metrics defined above and compare it with the baseline models. All the models are trained on 30,000 videos from Vox-KP for 200 epochs with hyper-parameters' default value for a fair comparison. The results for quantitative comparison are shown in the Table~\ref{tab:table1}. We can see the contribution of each of the components for preserving the facial structure of the target identity and having a temporally smooth keypoint sequence. Frame by frame comparison for all these models is also shown in the Figure~\ref{fig:compare}. Although Audio2Keypoint without discriminator performs slightly better in terms of L1 loss, this is justified as in the absence of the discriminator; the model learns to predict the keypoints' mean position mostly and hence low deviation from the ground truth. However, in the presence of discriminator, the model is pushed to generate a keypoint sequence that resembles the actual ground truth sequence, and hence the deviation as facial visemes form many-to-many mappings with the audio segments.

\begin{table}
  \begin{center}
    \caption{Quantitative Evaluation}
    \label{tab:table1}
     \begin{tabular}{l|c|c}
      \textbf{Models} & \textbf{Avg. L1} & \textbf{PCK}\\
      \hline
      Only Audio Input & 1.12 & 6.9\\
      Audio2Keypoint, without PIV encoder & 0.78 & 22.6\\
      Audio2Keypoint, without Discriminator & 0.56 & 23.21\\
      Audio2Keypoint, final model & 0.85 & 22.18\\ 
    \end{tabular}
  \end{center}
\end{table}

\subsection{Qualitative Analysis}
\label{sec:qualitative}

Audio2Keypointl can successfully generate a temporally smooth and coherent keypoint movement sequence, as shown in Figure~\ref{fig:first} and \ref{fig:compare-intra}. We qualitatively compare Audio2Keypoint synthesized keypoint sequences with baseline models in the Figure~\ref{fig:compare}. Audio2Keypoint focuses on the generation of the keypoint movement sequences and not on photo-realistic videos, we cannot compare with the existing methods. However, for visualization, we take the results from few of these models and compare with them in Figure~\ref{fig:compare}.

\section{Conclusion}
\label{sec:conclusion}

Human speech is often associated with facial expressions. However, the correlation between the two remains a complex map. This work creates an unique dataset, \emph{Vox-KP}, that has the facial keypoints movements associated with given audio segments. This dataset is used to for generating a plausible keypoint movement sequence in sync with the given audio. We see this work as a first step towards using facial keypoint movement to generate photo-realistic videos. Although, we learn to synthesize 2D facial keypoints, the model can also be adapted to learn 3D keypoints or to animate 3D head models by predicting each frame's parameters.

Currently, our method's key limitation is the lack of a keypoint guided image generation network \cite{zakharov2019few}, which prevents us from generating photo-realistic videos. Due to such a model's unavailability, we could not generate the video. Our immediate future work aims at synthesizing photo-realistic videos of a target person using his/her keypoint sequences generated using the proposed Audio2Keypoint model.


\begin{thebibliography}{10}\itemsep=-1pt

\bibitem{arik2018neural}
Sercan Arik, Jitong Chen, Kainan Peng, Wei Ping, and Yanqi Zhou.
\newblock Neural voice cloning with a few samples.
\newblock In {\em Advances in Neural Information Processing Systems}, pages
  10019--10029, 2018.

\bibitem{bulat2017far}
Adrian Bulat and Georgios Tzimiropoulos.
\newblock How far are we from solving the 2d \& 3d face alignment problem?(and
  a dataset of 230,000 3d facial landmarks).
\newblock In {\em Proceedings of the IEEE International Conference on Computer
  Vision}, pages 1021--1030, 2017.

\bibitem{cassell1999speech}
Justine Cassell, David McNeill, and Karl-Erik McCullough.
\newblock Speech-gesture mismatches: Evidence for one underlying representation
  of linguistic and nonlinguistic information.
\newblock {\em Pragmatics \& cognition}, 7(1):1--34, 1999.

\bibitem{chan2019everybody}
Caroline Chan, Shiry Ginosar, Tinghui Zhou, and Alexei~A Efros.
\newblock Everybody dance now.
\newblock In {\em Proceedings of the IEEE International Conference on Computer
  Vision}, pages 5933--5942, 2019.

\bibitem{chen2019hierarchical}
Lele Chen, Ross~K Maddox, Zhiyao Duan, and Chenliang Xu.
\newblock Hierarchical cross-modal talking face generation with dynamic
  pixel-wise loss.
\newblock In {\em Proceedings of the IEEE Conference on Computer Vision and
  Pattern Recognition}, pages 7832--7841, 2019.

\bibitem{chung2017you}
Joon~Son Chung, Amir Jamaludin, and Andrew Zisserman.
\newblock You said that?
\newblock {\em arXiv preprint arXiv:1705.02966}, 2017.

\bibitem{chung2018voxceleb2}
Joon~Son Chung, Arsha Nagrani, and Andrew Zisserman.
\newblock Voxceleb2: Deep speaker recognition.
\newblock {\em arXiv preprint arXiv:1806.05622}, 2018.

\bibitem{cootes1998active}
Timothy~F Cootes, Gareth~J Edwards, and Christopher~J Taylor.
\newblock Active appearance models.
\newblock In {\em European conference on computer vision}, pages 484--498.
  Springer, 1998.

\bibitem{gansynth}
Jesse Engel, Kumar~Krishna Agrawal, Shuo Chen, Ishaan Gulrajani, Chris Donahue,
  and Adam Roberts.
\newblock Gansynth: Adversarial neural audio synthesis.
\newblock 2019.

\bibitem{fan2015photo}
Bo Fan, Lijuan Wang, Frank~K Soong, and Lei Xie.
\newblock Photo-real talking head with deep bidirectional lstm.
\newblock In {\em 2015 IEEE International Conference on Acoustics, Speech and
  Signal Processing (ICASSP)}, pages 4884--4888. IEEE, 2015.

\bibitem{geng2018warp}
Jiahao Geng, Tianjia Shao, Youyi Zheng, Yanlin Weng, and Kun Zhou.
\newblock Warp-guided gans for single-photo facial animation.
\newblock {\em ACM Transactions on Graphics (TOG)}, 37(6):1--12, 2018.

\bibitem{ginosar2019learning}
Shiry Ginosar, Amir Bar, Gefen Kohavi, Caroline Chan, Andrew Owens, and
  Jitendra Malik.
\newblock Learning individual styles of conversational gesture.
\newblock In {\em Proceedings of the IEEE Conference on Computer Vision and
  Pattern Recognition}, pages 3497--3506, 2019.

\bibitem{isola2017image}
Phillip Isola, Jun-Yan Zhu, Tinghui Zhou, and Alexei~A Efros.
\newblock Image-to-image translation with conditional adversarial networks.
\newblock In {\em Proceedings of the IEEE conference on computer vision and
  pattern recognition}, pages 1125--1134, 2017.

\bibitem{kim2019neural}
Hyeongwoo Kim, Mohamed Elgharib, Michael Zollh{\"o}fer, Hans-Peter Seidel,
  Thabo Beeler, Christian Richardt, and Christian Theobalt.
\newblock Neural style-preserving visual dubbing.
\newblock {\em ACM Transactions on Graphics (TOG)}, 38(6):1--13, 2019.

\bibitem{kim2018deep}
Hyeongwoo Kim, Pablo Garrido, Ayush Tewari, Weipeng Xu, Justus Thies, Matthias
  Niessner, Patrick P{\'e}rez, Christian Richardt, Michael Zollh{\"o}fer, and
  Christian Theobalt.
\newblock Deep video portraits.
\newblock {\em ACM Transactions on Graphics (TOG)}, 37(4):1--14, 2018.

\bibitem{kingma2014adam}
Diederik~P Kingma and Jimmy Ba.
\newblock Adam: A method for stochastic optimization.
\newblock {\em arXiv preprint arXiv:1412.6980}, 2014.

\bibitem{kumar2019melgan}
Kundan Kumar, Rithesh Kumar, Thibault de Boissiere, Lucas Gestin, Wei~Zhen
  Teoh, Jose Sotelo, Alexandre de Br{\'e}bisson, Yoshua Bengio, and Aaron~C
  Courville.
\newblock Melgan: Generative adversarial networks for conditional waveform
  synthesis.
\newblock In {\em Advances in Neural Information Processing Systems}, pages
  14910--14921, 2019.

\bibitem{mirza2014conditional}
Mehdi Mirza and Simon Osindero.
\newblock Conditional generative adversarial nets.
\newblock {\em arXiv preprint arXiv:1411.1784}, 2014.

\bibitem{mori2012uncanny}
Masahiro Mori, Karl~F MacDorman, and Norri Kageki.
\newblock The uncanny valley [from the field].
\newblock {\em IEEE Robotics \& Automation Magazine}, 19(2):98--100, 2012.

\bibitem{nagano2018pagan}
Koki Nagano, Jaewoo Seo, Jun Xing, Lingyu Wei, Zimo Li, Shunsuke Saito, Aviral
  Agarwal, Jens Fursund, and Hao Li.
\newblock pagan: real-time avatars using dynamic textures.
\newblock {\em ACM Transactions on Graphics (TOG)}, 37(6):1--12, 2018.

\bibitem{nazzaro1970auditory}
James~R Nazzaro and Jean~N Nazzaro.
\newblock Auditory versus visual learning of temporal patterns.
\newblock {\em Journal of Experimental Psychology}, 84(3):477, 1970.

\bibitem{o1993conversations}
Brid O'Conaill, Steve Whittaker, and Sylvia Wilbur.
\newblock Conversations over video conferences: An evaluation of the spoken
  aspects of video-mediated communication.
\newblock {\em Human-computer interaction}, 8(4):389--428, 1993.

\bibitem{peng2017reconstruction}
Xi Peng, Xiang Yu, Kihyuk Sohn, Dimitris~N Metaxas, and Manmohan Chandraker.
\newblock Reconstruction-based disentanglement for pose-invariant face
  recognition.
\newblock In {\em Proceedings of the IEEE international conference on computer
  vision}, pages 1623--1632, 2017.

\bibitem{qian2019make}
Shengju Qian, Kwan-Yee Lin, Wayne Wu, Yangxiaokang Liu, Quan Wang, Fumin Shen,
  Chen Qian, and Ran He.
\newblock Make a face: Towards arbitrary high fidelity face manipulation.
\newblock In {\em Proceedings of the IEEE International Conference on Computer
  Vision}, pages 10033--10042, 2019.

\bibitem{ronneberger2015u}
Olaf Ronneberger, Philipp Fischer, and Thomas Brox.
\newblock U-net: Convolutional networks for biomedical image segmentation.
\newblock In {\em International Conference on Medical image computing and
  computer-assisted intervention}, pages 234--241. Springer, 2015.

\bibitem{song2020everybody}
Linsen Song, Wayne Wu, Chen Qian, Ran He, and Chen~Change Loy.
\newblock Everybody's talkin': Let me talk as you want.
\newblock {\em arXiv preprint arXiv:2001.05201}, 2020.

\bibitem{song2018talking}
Yang Song, Jingwen Zhu, Dawei Li, Xiaolong Wang, and Hairong Qi.
\newblock Talking face generation by conditional recurrent adversarial network.
\newblock {\em arXiv preprint arXiv:1804.04786}, 2018.

\bibitem{stacey2016contribution}
Paula~C Stacey, P{\'a}draig~T Kitterick, Saffron~D Morris, and Christian~J
  Sumner.
\newblock The contribution of visual information to the perception of speech in
  noise with and without informative temporal fine structure.
\newblock {\em Hearing research}, 336:17--28, 2016.

\bibitem{suwajanakorn2017synthesizing}
Supasorn Suwajanakorn, Steven~M Seitz, and Ira Kemelmacher-Shlizerman.
\newblock Synthesizing obama: learning lip sync from audio.
\newblock {\em ACM Transactions on Graphics (TOG)}, 36(4):1--13, 2017.

\bibitem{tang2019cycle}
Hao Tang, Dan Xu, Gaowen Liu, Wei Wang, Nicu Sebe, and Yan Yan.
\newblock Cycle in cycle generative adversarial networks for keypoint-guided
  image generation.
\newblock In {\em Proceedings of the 27th ACM International Conference on
  Multimedia}, pages 2052--2060, 2019.

\bibitem{tarasuik2013seeing}
Joanne Catherine~Ms Tarasuik, Jordy Kaufman, and Roslyn Galligan.
\newblock Seeing is believing but is hearing? comparing audio and video
  communication for young children.
\newblock {\em Frontiers in psychology}, 4:64, 2013.

\bibitem{taylor2016audio}
Sarah Taylor, Akihiro Kato, Ben Milner, and Iain Matthews.
\newblock Audio-to-visual speech conversion using deep neural networks.
\newblock 2016.

\bibitem{thies2019neural}
Justus Thies, Mohamed Elgharib, Ayush Tewari, Christian Theobalt, and Matthias
  Nie{\ss}ner.
\newblock Neural voice puppetry: Audio-driven facial reenactment.
\newblock {\em arXiv preprint arXiv:1912.05566}, 2019.

\bibitem{thies2019deferred}
Justus Thies, Michael Zollh{\"o}fer, and Matthias Nie{\ss}ner.
\newblock Deferred neural rendering: Image synthesis using neural textures.
\newblock {\em ACM Transactions on Graphics (TOG)}, 38(4):1--12, 2019.

\bibitem{thies2016face2face}
Justus Thies, Michael Zollhofer, Marc Stamminger, Christian Theobalt, and
  Matthias Nie{\ss}ner.
\newblock Face2face: Real-time face capture and reenactment of rgb videos.
\newblock In {\em Proceedings of the IEEE conference on computer vision and
  pattern recognition}, pages 2387--2395, 2016.

\bibitem{vougioukas2018end}
Konstantinos Vougioukas, Stavros Petridis, and Maja Pantic.
\newblock End-to-end speech-driven facial animation with temporal gans.
\newblock {\em arXiv preprint arXiv:1805.09313}, 2018.

\bibitem{vougioukas2019realistic}
Konstantinos Vougioukas, Stavros Petridis, and Maja Pantic.
\newblock Realistic speech-driven facial animation with gans.
\newblock {\em International Journal of Computer Vision}, pages 1--16, 2019.

\bibitem{wiles2018x2face}
Olivia Wiles, A Sophia~Koepke, and Andrew Zisserman.
\newblock X2face: A network for controlling face generation using images,
  audio, and pose codes.
\newblock In {\em Proceedings of the European conference on computer vision
  (ECCV)}, pages 670--686, 2018.

\bibitem{yang2012articulated}
Yi Yang and Deva Ramanan.
\newblock Articulated human detection with flexible mixtures of parts.
\newblock {\em IEEE transactions on pattern analysis and machine intelligence},
  35(12):2878--2890, 2012.

\bibitem{yehia1998quantitative}
Hani Yehia, Philip Rubin, and Eric Vatikiotis-Bateson.
\newblock Quantitative association of vocal-tract and facial behavior.
\newblock {\em Speech Communication}, 26(1-2):23--43, 1998.

\bibitem{yin2017multi}
Xi Yin and Xiaoming Liu.
\newblock Multi-task convolutional neural network for pose-invariant face
  recognition.
\newblock {\em IEEE Transactions on Image Processing}, 27(2):964--975, 2017.

\bibitem{zakharov2019few}
Egor Zakharov, Aliaksandra Shysheya, Egor Burkov, and Victor Lempitsky.
\newblock Few-shot adversarial learning of realistic neural talking head
  models.
\newblock In {\em Proceedings of the IEEE International Conference on Computer
  Vision}, pages 9459--9468, 2019.

\bibitem{zhou2019talking}
Hang Zhou, Yu Liu, Ziwei Liu, Ping Luo, and Xiaogang Wang.
\newblock Talking face generation by adversarially disentangled audio-visual
  representation.
\newblock In {\em Proceedings of the AAAI Conference on Artificial
  Intelligence}, volume~33, pages 9299--9306, 2019.

\bibitem{zhu2018high}
Hao Zhu, Aihua Zheng, Huaibo Huang, and Ran He.
\newblock High-resolution talking face generation via mutual information
  approximation.
\newblock {\em arXiv preprint arXiv:1812.06589}, 2018.

\bibitem{zollhofer2018state}
Michael Zollh{\"o}fer, Justus Thies, Pablo Garrido, Derek Bradley, Thabo
  Beeler, Patrick P{\'e}rez, Marc Stamminger, Matthias Nie{\ss}ner, and
  Christian Theobalt.
\newblock State of the art on monocular 3d face reconstruction, tracking, and
  applications.
\newblock In {\em Computer Graphics Forum}, volume~37, pages 523--550. Wiley
  Online Library, 2018.

\end{thebibliography}
\end{document}